\documentclass[twocolumn]{entalpic}

\usepackage{amsmath}
\usepackage{amssymb}
\usepackage{mathtools}
\usepackage[numbers,sort&compress]{natbib}
\usepackage{array}
\usepackage{url}
\usepackage{ragged2e}
\usepackage{algorithm}
\usepackage{algpseudocode}

\newcolumntype{L}[1]{>{\RaggedRight\arraybackslash}p{#1}}

\title{Keeping the Agent Out of the Hot Loop: Self-Healing DFT Workflows Across HPC Centers}

\author[1,*]{Daniel T. Speckhard}
\author[1]{Lucas Pinede}
\author[1,*]{Sagar Pal}
\author[1]{Ali Ramlaoui}
\author[1]{Hannah Bull}
\author[1]{Cory Hargus}
\author[1]{Victor Schmidt}
\author[1,*]{Alexandre Duval}

\affiliation[1]{Entalpic, Paris, France}
\contribution[*]{Corresponding authors}
\metadata[Correspondence]{\email{\{daniel.speckhard, sagar.pal, alexandre.duval\}@entalpic.ai}}

\abstract{
High-throughput Density Functional Theory (DFT) workflows are widely used in computational chemistry, materials science, and condensed matter physics, but large-scale runs still require substantial human intervention when calculations, workflow daemons, queues, or archival steps fail. LLM agents can help reduce this supervision, but our deployment experience shows that giving them direct control to launch, cancel, or modify jobs risks wasting expensive compute or human debugging time. Motivated by this, we introduce LeDFTBuddy, an agentic monitoring and build workflow that helps construct a deterministic self-healing control plane while keeping LLM agents outside the production hot path. LeDFTBuddy observes the workflow state through typed read-only tools, diagnoses unresolved incidents, writes issues and draft PRs, and routes proposed fixes through adversarial review and human approval before deployment. The multi-agent workflow helped build and harden a deterministic outer loop for common DFT workflow failures and provide a queue migration policy across computational infrastructure for stalled jobs. Across multiple hardware partitions and HPC centers, the outer loop observed 1,677 remote workflow errors, recovered 731 of them with deployed logic, backfilled 8,343 missing records, and ran 15 queue migrations without duplicate live executions. The main lesson is that agents are useful for finding missing policies, but state-changing recovery for large-scale scientific workflows should remain deterministic or human-approved.
}

\begin{document}

\maketitle

\section{Introduction}

Density functional theory (DFT) is a quantum-mechanical simulation method used to predict the behavior of molecules and materials~\cite{hohenberg1964inhomogeneous,giustino2014materials}. Applications range from catalyst screening and molecular stability prediction to battery materials, alloys, and thermal transport~\cite{carbogno2017ab,rosen2019identifying, nair2025materials}. In high-throughput DFT, practitioners run many related calculations~\cite{petousis2017high}. These jobs are usually run on shared high-performance computing (HPC) systems with different CPU and GPU queues, scheduler policies, software environments, allocations, and queue states~\cite{gavini2023roadmap,speckhard2025workflows}. The workflow goal is therefore not only to run one simulation, but to keep a distributed execution of multi-step calculations consistent while jobs, daemons, queues, databases, and archives change state.

An ecosystem of software tools makes parts of this process more robust and automated~\cite{leser2026workflow}. Libraries such as ASE~\cite{ase-paper} and pymatgen~\cite{ong2013python} help users create and manipulate atomistic systems programmatically. Workflow tools such as jobflow~\cite{rosen2024jobflow} represent multi-step calculation graphs, while jobflow-remote submits those jobs to HPC schedulers and records their remote state. At the calculation level, Custodian~\cite{ong2013python,jain2013commentary} monitors output from the DFT code VASP~\cite{kresse1996efficiency} for known error signatures and applies deterministic fixes and restarts such as changing numerical parameters affecting the calculation's convergence. These tools remove a large amount of manual setup time and supervision, but large-scale executions still require users to choose parameters, monitor queues and convergence, and to restart and adjust many failed calculations. Failures can come from the DFT calculations themselves, but also from HPC queue congestion, dropped remote connections to the workflow manager, compute time allocation exhaustion, post-processing archival failures, and expiring scratch storage.

\begin{figure*}[!t]
\centering
\includegraphics[width=\textwidth]{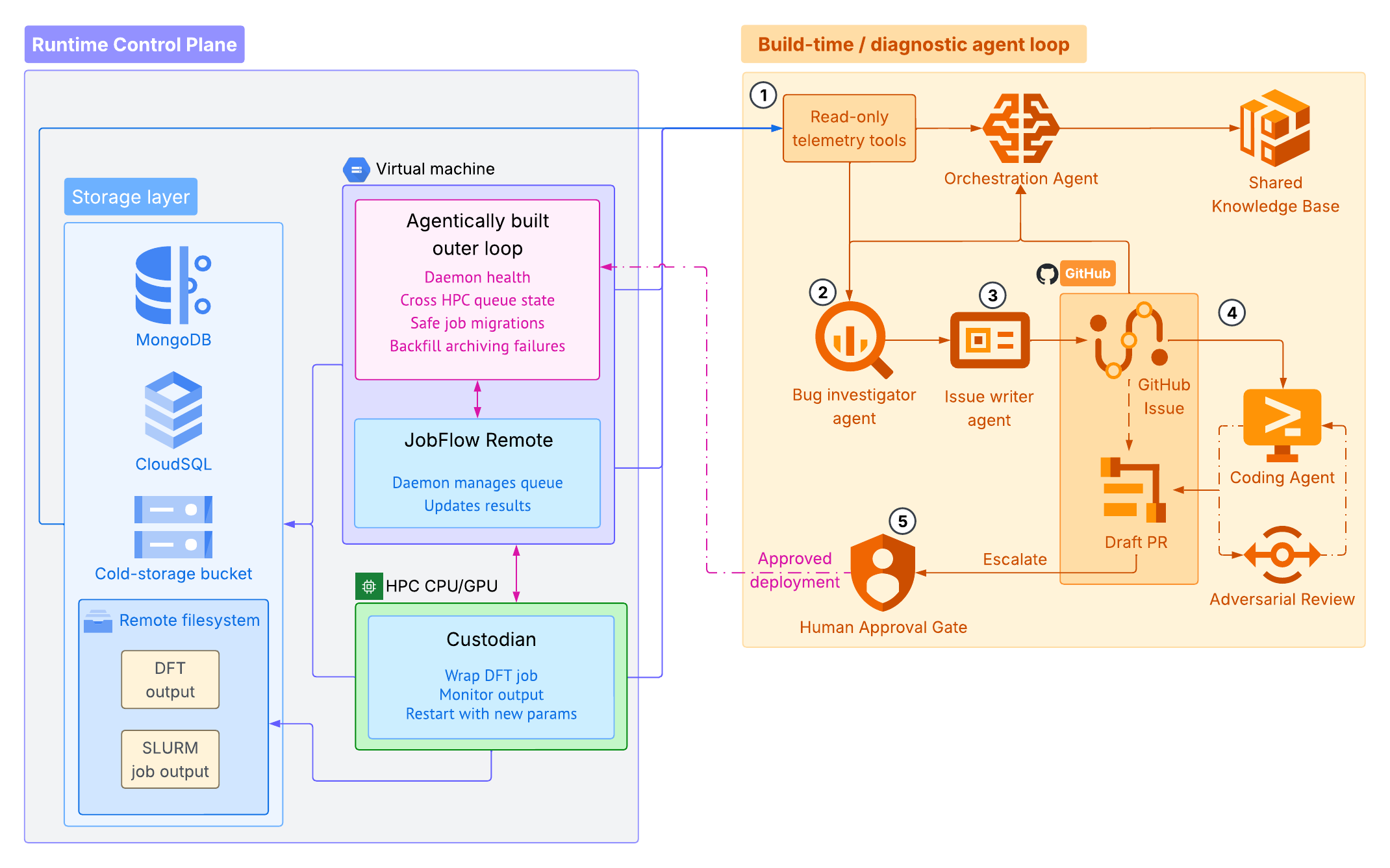}
\caption{Boundary between deterministic runtime recovery and agent-assisted build-time hardening. On the runtime side, Custodian handles known VASP failures, jobflow-remote submits and tracks jobs, and a DBOS-backed outer loop, which LeDFTBuddy built, reconciles runner health, queue state, archive completeness, and migration safety. On the build-time side, LeDFTBuddy observes state through read-only telemetry tools. Unresolved or recurring failures become issues, draft PRs, adversarial reviews, and human-approved updates to the deterministic outer loop.}
\label{fig:three_loop_architecture}
\end{figure*}

Recent work has explored LLM-based agents for computational chemistry and DFT workflows, including VASPilot~\cite{liu2025vaspilot}, DREAMS~\cite{wang2025dreams}, ChemGraph~\cite{pham2026chemgraph}, ChemCrow~\cite{bran2023chemcrow}, and Coscientist~\cite{boiko2023autonomous}. These LLM-assisted workflows demonstrate that agents can call tools, generate inputs, inspect errors, query documentation, summarize outputs, and coordinate multi-step scientific calculations. Prior systems mainly ask how much of a scientific workflow an agent can automate. We ask a slightly different question. \textit{Namely, where should the agent sit in a production DFT campaign?} This matters because some actions are not just wrong, but are destructive and impossible to undo. In our experience, a stochastic agent with direct control over execution can duplicate submissions, kill the wrong job, move jobs unnecessarily, miss archival failures, or apply unreviewed code fixes. These mistakes are costly for executions of large-scale HPC workflows, where the wrong action can waste compute allocations and human debugging time, and make the workflow state harder to recover. This is a familiar control-plane principle in production systems, but LLM agents make the boundary important for scientific workflows that expose expensive jobs,  compute hardware, scheduler state, databases, filesystems, and code~\cite{dawson2026lara, doshi2026towards}. Our design therefore keeps the runtime loop automated but deterministic, and uses LLM-based agents only for read-only diagnosis, failure summarization, issue writing, draft PRs, and human-gated proposed fixes when deterministic rules cause or fail to prevent unforeseen failure modes.

This paper makes three main contributions.

\begin{itemize}
    \item We report deployment lessons from giving agents access to parts of a production DFT workflow across CPU and GPU partitions on two different HPC centers. The failures show why the agent should not be given access to state-changing actions on the workflow hot path.
    \item We present an agent monitoring and build workflow that uses typed read-only telemetry, issue writing, draft PRs, adversarial review, and human approval to turn recurring failures into reviewed code changes.
    \item We present two case studies showing how the investigative and build workflow helps turn recurring failures into reusable deterministic workflow infrastructure. The first is a durable outer loop for runner restarts, remote-error recovery, archive backfills, and environment checks. The second is a queue-migration policy for jobs stalled by exhausted allocations, low fair-share priority, or maintenance related outages.

\end{itemize}

\section{Experience Report from Direct Agent Control}

We began with a high-throughput workflow for multi-step DFT calculations using the VASP program~\cite{kresse1996efficiency}. We make use of atomate2~\cite{ganose2025atomate2} workflow recipes and submit jobs through jobflow-remote to GENCI's Jean Zay and Adastra HPC centers, spanning CPU and GPU partitions under two compute allocations. The VASP jobs are wrapped with Custodian, which handles known calculation-level failures through predefined fixes and restarts. In our deployment, workflow state was stored in MongoDB, parsed DFT results were written to CloudSQL, and completed job directories were archived to cloud storage.

The failures we observed were not only DFT failures. There were state-reconciliation failures across workflow state, scheduler state, database state, archives, and code state. Table~\ref{tab:failure-modes} groups the main failure modes by the bounded response they motivated, including calculation failures that exhausted Custodian, \texttt{REMOTE\_ERROR} states in jobflow-remote, stalled or poorly placed queued jobs, missing parsed rows, and cross-host environment drift where different Python environments wrote or read incompatible schemas. Of 12 documented human-reviewed escalations, 7 required DFT-specific diagnosis, including input-parameter mistakes, restart behavior, schema drift, output-state interpretation, and slab-specific corrections.

We then tried giving an LLM-based agent direct control over parts of this workflow. This was useful for investigation, but brittle for production actions. With skills or MCP tools, the agent could retrieve information from MongoDB, CloudSQL, cloud archives, and HPC logs more reliably. 

Without these typed tools, the agent often spent long loops debugging database, archive, or HPC access. With these typed tools, it could inspect state more reliably, but it still made unsafe choices when allowed to act directly. It killed jobs unnecessarily, restarted jobs without canceling the old ones, or proposed code fixes that only handled part of the failure. These mistakes cost compute time ($\approx$50,000 CPU hours, which we detail further below) and in many cases created extra debugging work for the human to restore the workflow state. This experience led to the boundary used in the rest of the paper. Agents may investigate and propose fixes, while production state changes follow deterministic policies or human approval.
\section{LeDFTBuddy: Agent Monitoring and Build Workflow}
The direct-control failures changed how we used the agent. We moved the agent out of the hot path and into \textsc{LeDFTBuddy}, a diagnostic and build-time agent workflow to help build determinism into the large-scale scientific workflow. Figure~\ref{fig:three_loop_architecture} shows this boundary and describes how we use multiple agents, each given a bounded role. The left side is the runtime control plane. It contains the existing Custodian and jobflow-remote loops, plus a durable outer loop that LeDFTBuddy helped build and harden. The right side of the figure shows the build-time agent loop. It observes selected runtime state through read-only telemetry tools, produces issues, draft PRs, reviews, and human-approved changes. Each PR was reviewed by two adversarial reviewers using different model families, Claude Sonnet 4.5 and GPT-5.5. A human still gates the decision to accept and deploy the change. Across the current LeDFTBuddy workflow, 35 PRs were merged. One required significant human intervention, and two later required revert or post-deployment fixes. For each escalated incident, LeDFTBuddy records the typed-tool evidence, agent diagnosis, reviewer comments, human decision, linked issue or PR, and post-deployment outcome.

An orchestration agent manages the different agents. It observes MongoDB state, CloudSQL results, cloud archives, and selected DFT and Slurm logs through read-only telemetry tools. These tools keep context small and avoid broad database, filesystem, or scheduler access. When the loops reach an ambiguous state, the orchestrator dispatches the build-time agent workflow described above. The agents can investigate the failure, write an issue, propose a draft PR, and route it through adversarial review. In a 20-query benchmark, typed tools returned 596 tokens compared with 8,913 for scoped raw queries, while both answered all queries on the first attempt. In real campaign transcripts, only 62.6\% of 91 ad hoc database or filesystem attempts succeeded on the first attempt. This is why the agent reads state through typed tools rather than writing its own production queries.

\begin{table*}[!t]
\centering
\caption{Observed failures and the bounded responses added during deployment.}
\label{tab:failure-modes}
\footnotesize
\setlength{\tabcolsep}{4pt}
{\renewcommand{\arraystretch}{1.14}
\begin{tabular}{@{}L{0.20\textwidth} L{0.38\textwidth} L{0.36\textwidth}@{}}
\toprule
\textbf{Observed issue} & \textbf{Bounded response} & \textbf{Deployment evidence} \\
\midrule
\addlinespace[0.30em]
Runner loss &
Reconnect or restart jobflow-remote after missed heartbeats &
25 failures over 76 days. Median recovery 21 seconds. 95th percentile recovered within 116 seconds\\
\addlinespace[0.30em]
\texttt{REMOTE\_ERROR} state &
Classify remote failures, rerun supported cases, or escalate &
1,677 jobs entered \texttt{REMOTE\_ERROR}. 731 recovered. Remaining cases triaged \\
\addlinespace[0.30em]
CloudSQL persistence gap &
Backfill missing parsed rows from archived outputs &
8,343 rows backfilled \\
\addlinespace[0.30em]
Cross-host environment drift &
Pin environments and add schema checks across hosts &
69 archive jobs broken. 483 validation errors \\
\addlinespace[0.30em]
Ad hoc DB inspection &
Replace ad hoc CloudSQL and MongoDB scripts with typed read-only tools &
20 queries. 596 tokens with typed tools vs. 8,913 for scoped raw queries. 91 ad hoc attempts had 62.6\% first-attempt success \\
\addlinespace[0.30em]
DFT/config escalations &
Collect scheduler, workflow, and DFT evidence for human-reviewed fixes &
12 incidents. 7 DFT-specific. Median detection $\sim$1 d. Two outliers took 8--10 d \\
\addlinespace[0.30em]
Custodian retries exhausted &
Scan retry archives and escalate VASP jobs that exhausted built-in fixes &
29,296 run directories scanned. 294 exhausted retries. Physics-specific convergence failures dominated\\
\addlinespace[0.30em]
Direct job migration by agent &
Move state-changing actions out of the agent hot path &
Direct migration failed in 2 of 5 attempts and wasted roughly 50,000 CPU hours \\
\addlinespace[0.30em]
Queue migration policy &
Use deterministic thresholds plus cancel, confirm, resubmit, and exactly-one-live-job checks &
15 deployed migrations. No duplicate live executions observed \\

\bottomrule
\end{tabular}}
\end{table*}

\section{Case Study 1: the outer loop}

After deploying LeDFTBuddy, the first recurring issues were durability failures around the Custodian and jobflow-remote loops. Custodian can, however, encounter failures that it does not automatically address. Jobflow-remote submits jobs to the HPC remotely via SSH connections. Job states, such as awaiting submission or running, are stored in a local MongoDB database. This helps circumvent a typical HPC constraint that prohibits processes from running indefinitely. LeDFTBuddy surfaced lost daemon SSH connections, stalled daemon processes, failed post-processing archive jobs, and Python virtual-environment mismatches introduced by deployed fixes.

To address these issues, LeDFTBuddy proposed, in successive PRs, an outer loop to monitor the health of the DFT jobs, Custodian restarts, jobflow-remote daemon, archiving jobs and HPC health and queue status checks. The outer loop is implemented with DBOS~\cite{skiadopoulos2021dbos}, which persists recovery tasks and their intermediate state so runner restarts, backfills, and migration checks can resume safely after local failures. The outer loop can reconnect or restart a failed jobflow-remote runner, rerun \texttt{REMOTE\_ERROR} cases, backfill missing parsed metadata or archived files into the storage layer, and run environment and schema checks across hosts. It can also scan Custodian retry archives and escalate VASP jobs that exhausted built-in fixes. Among 294 Custodian-exhausted run directories, the dominant failures were physics-specific configuration or convergence failures rather than generic infrastructure errors. Ambiguous states are not repaired directly. They are surfaced to LeDFTBuddy or a human reviewer.

Table~\ref{tab:failure-modes} shows the operational scale of these cases during deployment. Most runner recoveries took well under a minute. The data-plane checks backfilled 8,343 missing CloudSQL rows from archived outputs. Environment checks found 69 broken archive jobs and 483 validation errors. These results show the more limited pattern we want. Once a failure mode is understood, the state-changing response can move from an agent-proposed fix into a bounded deterministic recovery path.

\section{Case Study 2: Queue Migration Policy}

HPC queues are often congested, and queue wait can dominate workflow placement decisions~\cite{smith1999using,nurmi2006evaluation}. Large-scale executions can span allocations, partitions, and HPC centers with different wait times~\cite{alsaadi2022radical,da2023workflows}. This case study arose from recurring stalled jobs surfaced by LeDFTBuddy and addressed through a deterministic policy in the outer loop.

Some jobs stopped making progress because an HPC partition underwent maintenance. In some cases, maintenance lasted more than 24 hours. During a heatwave, a GPU partition was shut for 6 days. In other cases, allocations had expired, been spent, or fair-share was too low for the queue to be useful. Moving jobs from one HPC center to another directly with an agent was unsafe. In five early direct-migration attempts, two left the old queued jobs in place after submitting replacement jobs. Slurm-based accounting for those duplicate calculations amounted to roughly 50,000 CPU hours.

After LeDFTBuddy was deployed, the investigator agent reported stalled jobs and the target HPC as unreachable. LeDFTBuddy proposed a queue migration policy to move jobs from one HPC center to another. Human review made the policy more conservative by increasing uncertainty penalties and treating unreachable targets as likely down for maintenance rather than assuming a quick recovery. LeDFTBuddy drafted PRs for an outer-loop deterministic policy spanning allocations, queues, and HPCs. For each queued job, the outer loop estimates time to completion on each compatible target $c$:
\begin{equation}
\label{eq:queue_migration_equation}
\widehat{T}_c = W_c + R_c + M_c + U_c,
\end{equation}
\noindent where $W_c$ is the estimated wait time on target $c$, $R_c$ is the expected runtime on that target, $M_c$ is migration overhead, and $U_c$ is an uncertainty penalty. The policy asks whether the expected improvement is large enough to justify giving up the current queue position. A job moves from its current target $c_0$ to the best target $c^\star$ only when
\begin{equation}
\widehat{T}_{c_0} - \widehat{T}_{c^\star} > \max(\Delta_{\min}, \lambda W_{c_0} + m),
\end{equation}
\noindent and the job is in a safe-to-move state. 
Here, $\lambda W_{c_0}$ scales the required improvement with the estimated remaining wait on the current target, so migration occurs when another target offers a sufficiently large improvement relative to that wait. $m$ is an anti-churn margin, and $\Delta_{\min}$ is the minimum absolute improvement. For the deployment reported here, the policy used $\Delta_{\min}=1$ hour (h), $\lambda=0.5$, and $m=0.5$ h, so a job with a current estimated wait of 10 h requires a 5.5 h improvement, while a job with a 48 h wait requires a 24.5 h improvement before a migration occurs.

A target is compatible only if it is reachable and satisfies device, allocation, walltime, and queue-health checks. Migration uses a deterministic cancel-confirm-resubmit protocol. The outer loop marks the job as migration-pending, confirms cancellation or checkpointing, resubmits, and verifies that exactly one live execution exists. Ambiguous states are escalated to a human. LeDFTBuddy monitors migrations, but it does not execute them. Across 15 deployed migrations, no duplicate live executions were observed. In one representative migration, jobs moved from the Adastra HPC cluster to Jean Zay finished within two days, while a probe job left in the original Adastra queue waited eight days before starting, yielding at least a six-day completion-time improvement.

\section{Limitations and Conclusion}

We introduced LeDFTBuddy, a human-gated agent monitoring and build workflow for large-scale DFT simulations. The central design choice is to keep the agent outside the runtime hot path. This design was motivated by costly failures from direct agent control.

LeDFTBuddy helped build reusable components for more durable workflows, including typed telemetry tools, a DBOS-based outer loop for campaign supervision, and a queue migration policy for stalled jobs. These components can be used without running the agentic workflow in production. Although our deployment is in DFT, the same boundary may be useful for other simulation programs and large-scale science workflows where agents interact with expensive shared infrastructure. The broader lesson is that LLM-based agents should not replace large-scale scientific workflow engines. For high-throughput DFT, they are currently more useful as monitors, investigators, and software-development assistants around those engines.

\section*{Acknowledgements}

The authors used OpenAI Codex, Claude, and Gemini to assist with drafting, editing, and figure-generation code. All benchmark results, implementation details, and conclusions are the authors' own. We thank Richard Tran and Martin Siron for their valuable input. This work was performed using HPC resources from GENCI-IDRIS and GENCI-CINES under the allocations 2025-A0181016212, 2025-AD011016353, and 2026-A0201016212 on the Jean Zay and Adastra supercomputers.

\bibliographystyle{plainnat}
\bibliography{references}

\clearpage
\appendix
\raggedbottom

\section{Custodian failure breakdown}
\label{app:custodian_failures}

Table~\ref{tab:custodian_handlers} reports the handlers observed across the 294 run directories that exhausted retries. Counts exceed the number of run directories because several handlers can fire during one run. Table~\ref{tab:vasp_error_signatures} gives the dominant VASP signatures.

\begin{table}[H]
\centering
\caption{Custodian handlers observed in exhausted VASP runs.}
\label{tab:custodian_handlers}
\footnotesize
\setlength{\tabcolsep}{5pt}
\renewcommand{\arraystretch}{1.1}
\begin{tabular}{@{}L{0.58\linewidth}rr@{}}
\toprule
\textbf{Handler} & \textbf{Calls} & \textbf{Run dirs} \\
\midrule
\texttt{VaspErrorHandler} & 490 & 247 \\
\texttt{UnconvergedErrorHandler} & 354 & 244 \\
\texttt{DelayedWalltimeHandler} & 164 & 130 \\
\texttt{PotimErrorHandler} & 10 & 4 \\
\texttt{PositiveEnergyErrorHandler} & 2 & 2 \\
\texttt{LargeSigmaHandler} & 1 & 1 \\
\texttt{NonConvergingErrorHandler} & 1 & 1 \\
\bottomrule
\end{tabular}
\end{table}

\begin{table}[H]
\centering
\caption{Dominant VASP signatures in exhausted Custodian runs.}
\label{tab:vasp_error_signatures}
\footnotesize
\setlength{\tabcolsep}{4pt}
\renewcommand{\arraystretch}{1.1}
\begin{tabular}{@{}L{0.25\linewidth}rL{0.55\linewidth}@{}}
\toprule
\textbf{Signature} & \textbf{Count} & \textbf{Interpretation} \\
\midrule
\texttt{brions} & 486 & Ionic-dynamics failure from high-force or unphysical geometry \\
Unconverged & 354 & Ionic relaxation reached \texttt{NSW} before force convergence \\
Walltime reached & 164 & \texttt{STOPCAR} written before scheduler walltime \\
\texttt{POTIM} & 10 & Ionic step size too large \\
\texttt{eddrmm} & 2 & Electronic-minimization failure \\
\texttt{brmix} & 2 & Charge-mixing failure \\
\texttt{subspacematrix} & 2 & Subspace-diagonalization failure \\
Positive energy & 2 & Structurally corrupt or unstable configuration \\
\texttt{LargeSigma} & 1 & Smearing width too large \\
Non-converging job & 1 & Repeated SCF non-convergence \\
\bottomrule
\end{tabular}
\end{table}

\section{Tool-access diagnostics}
\label{app:tool_diagnostics}

The controlled benchmark shows that typed tools reduce the required context size. The real incident traces in Table~\ref{tab:adhoc_query_failures} show why bounded interfaces were also more reliable than ad hoc agent-written queries.

\begin{table}[H]
\centering
\caption{Outcomes of 91 real ad hoc database and filesystem invocations.}
\label{tab:adhoc_query_failures}
\footnotesize
\setlength{\tabcolsep}{6pt}
\renewcommand{\arraystretch}{1.1}
\begin{tabular}{@{}lr@{}}
\toprule
\textbf{Outcome} & \textbf{Count} \\
\midrule
First-attempt success & 57 \\
Timeouts from over-fetching or unbounded scans & 17 \\
Syntax errors & 8 \\
Authentication issues & 3 \\
Wrong field path & 1 \\
Other failures & 5 \\
\bottomrule
\end{tabular}
\end{table}

\end{document}